\documentclass[letterpaper, 10 pt, conference]{ieeeconf}  
\IEEEoverridecommandlockouts
\usepackage{lipsum}
\usepackage{graphics}
\usepackage{epsfig}
\usepackage{mathptmx} 
\usepackage{amsmath}
\usepackage{amssymb}
\usepackage{balance}
\usepackage{multirow}
\usepackage{algorithmic}
\usepackage{algorithm}

\usepackage{graphicx}
\usepackage{textcomp}
\usepackage{xcolor}
\usepackage{balance}
\usepackage{subcaption}
\usepackage{verbatim}
\usepackage{dblfloatfix}
\usepackage{xcolor}
\usepackage{stackrel}
\usepackage{float}
\usepackage[colorlinks=black,linkcolor=black]{hyperref}

\usepackage{booktabs}
\usepackage{url}
\makeatletter
\newcommand\xleftrightarrow[2][]{%
  \ext@arrow 9999{\longleftrightarrowfill@}{#1}{#2}}
\newcommand\longleftrightarrowfill@{%
  \arrowfill@\leftarrow\relbar\rightarrow}
\makeatother

\title{ \LARGE \bf
    Graph-based Normalizing Flow for Human Motion \\   Generation and Reconstruction
}
\author{
    Wenjie Yin$^{1}$, 
    Hang Yin$^{1}$, 
    Danica Kragic$^{1}$, 
    Mårten Björkman$^{1}$
    \thanks{
        $^{1}$Robotics, Perception and Learning lab, KTH Royal Institute of Technology, Sweden.
        {\tt\small \{yinw, hyin, dani, celle\}@kth.se}.
    }
}
\begin{document}
\maketitle
\thispagestyle{empty}
\pagestyle{empty}
\begin{abstract}

Data-driven approaches for modeling human skeletal motion have found various applications in interactive media and social robotics. Challenges remain in these fields for generating high-fidelity samples and robustly reconstructing motion from imperfect input data, due to e.g. missed marker detection.
In this paper, we propose a probabilistic generative model to synthesize and reconstruct long horizon motion sequences conditioned on past information and control signals, such as the path along which an individual is moving. Our method adapts the existing work MoGlow by introducing a new graph-based model. 
The model leverages the spatial-temporal graph convolutional network (ST-GCN) to effectively capture the spatial structure and temporal correlation of skeletal motion data at multiple scales. 
We evaluate the models on a mixture of motion capture datasets of human locomotion with foot-step and bone-length analysis.
The results demonstrate the advantages of our model in reconstructing missing markers and achieving comparable results on generating realistic future poses. When the inputs are imperfect, our model shows improvements on robustness of generation. 

\end{abstract}

\section{INTRODUCTION}
Modeling human motion is essential and challenging in human-robot interaction. Computational models that capture rich motion patterns can facilitate animating synthetic characters \cite{henter2020moglow} and understanding human behaviors for greater autonomy in social robotics scenarios~\cite{sidebyside2014hri}.
Recent work \cite{fragkiadaki2015recurrent, butepage2017deep, martinez2017human} have proposed deterministic data-driven motion synthesis methods. 
These methods are often limited to generation of stereotypical samples and fail to capture the natural variability of human motion.
Probabilistic generative models, however, permit modelling of the full space of possible poses without collapsing to an average pose \cite{wang2019combining, ahn2020generative, habibie2017recurrent}. 
Normalizing flow based approaches have so far received less attention compared to the alternatives, and have rarely been explored
for human motion synthesis \cite{henter2020moglow}. 

One central challenge that generative models are facing is to achieve robust synthesis under imperfect conditions. 
Human skeletal motion is commonly collected using motion capture (MoCap) systems such as Vicon\footnotemark[2] or OptiTrack\footnotemark[3].
When moved to less controlled environments,
these systems inevitably suffer from missed markers due to occlusion or other detection failures \cite{kucherenko2018neural}. 
Unfortunately, most previous works do not yield satisfactory performance in generating stable and consistent motion patterns under such conditions. 

\begin{figure}[t]
  \centering
  \includegraphics[width=0.48\textwidth]{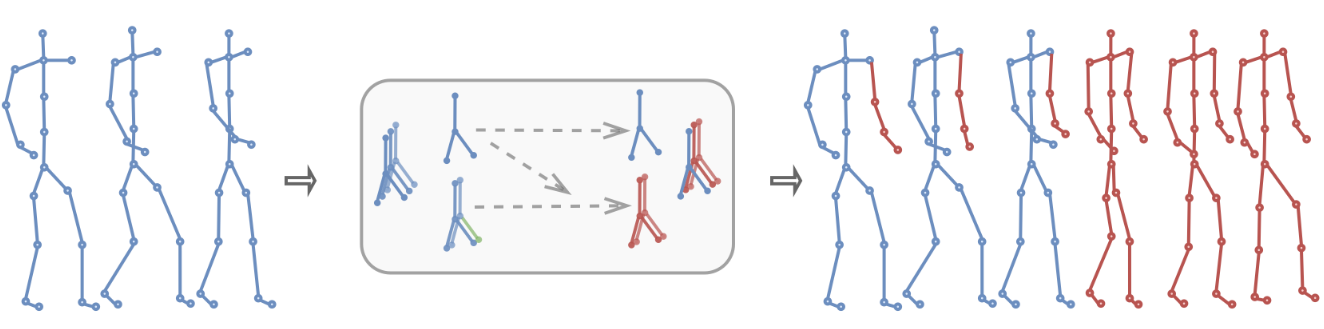}
  \caption{ 
    A probabilistic framework for human motion generation and reconstruction: Generating future human motion given partially observed past pose sequences using graph-based normalizing flow and reconstructing the missing markers in past sequences. 
  }
  \label{fig:front-page}
\end{figure}

We propose a graph-based probabilistic generative model to address this limitation in the context of human motion generation and reconstruction, as illustrated in Fig. \ref{fig:front-page}.
The model builds upon MoGlow \cite{henter2020moglow}, an autoregressive normalizing flow model. The idea is to exploit the invariant spatial correlation of human skeletons and to enforce this inductive bias with a graph structure.
A spatial configuration partition strategy orders the nodes in a spatial graph neighbor set, allowing for convolutions on graphs. 
By using graph convolutional networks, the model encodes correlations between markers and hence achieves a robust synthesis.
Our framework is designed as a two-step pipeline. 
The first step is to generate future poses given incomplete past sequences and control information.
The second step is to reconstruct the missing markers by reversing the generated sequences. 
We evaluate our framework on a mixture of datasets of human locomotion. 
The evaluation shows that, even with imperfect input data, our proposed graph motion glow model attains a reconstructed motion quality close to that of recorded motion capture data, outperforming a state-of-the-art baseline. 
\footnotetext[2]{\url{http://www.vicon.com/}}
\footnotetext[3]{\url{https://optitrack.com/}}

In summary, our work contributes
\begin{itemize}
    \item an integration  of graph neural network and normalizing flows to improve the encoding of human skeletal motion, and
    \item extending MoGlow to generate human motion patterns given incomplete past poses and managing to reconstruct missing markers robustly.
\end{itemize}

The paper proceeds as follows:
Section \ref{sec:related} gives an introduction to previous works on the topics of human motion generation and reconstruction, including graph-based human body representation. 
Section \ref{sec:method} formulates the problem and describes the full design of our proposed framework for motion generation and reconstruction. 
Section \ref{sec:experiment} presents the experimental setup and discusses results, compares the results to the baseline. 
Finally, section \ref{sec:result} outlines a plan for future work and concludes the paper. 
 
\section{RELATED WORK}
\label{sec:related}
In this section, 
we provide an overview of deep learning based motion generation (Section \ref{sec:related-dg}) 
and pose reconstruction methods from partially observable data (Section \ref{sec:related-mm}), 
and then we describe prior works that use graph-based human motion representation (Section \ref{sec:related-gb}).

\subsection{Deep Learning for Motion Prediction and Generation}
\label{sec:related-dg}
Deep learning approaches excel at processing massive amounts of data with no requirements on feature engineering. Such approaches have been widely adopted for human motion generation following earlier successes in other domains. 
Fragkiadaki et al. \cite{fragkiadaki2015recurrent} present an encoder-recurrent-decoder (ERD) to jointly learn a skeleton embedding with sequential information for human pose prediction. 
Butepage et al. \cite{butepage2017deep} propose a temporal encoding-decoding network to encode previous frames and decode future sequences. 
Martinez et al. \cite{martinez2017human} develop a sequence-to-sequence architecture with residual connections to predict joint velocities, while
Li et al. \cite{li2017auto} propose the auto-conditioned LSTM to synthesize complex and long-term motion patterns.

The approaches reviewed above predict deterministic samples for a given input.
To synthesize diverse motion patterns, one way is to use probabilistic generative approaches. 
Generative adversarial networks (GANs) and their variants have been applied in human motion synthesis and control \cite{yan2019convolutional, wang2019combining} and speech-driven motion generation \cite{sadoughi2018novel}. 
GANs are deemed more effective for encoding modes of the data but can be difficult to train and evaluate.
As an alternative, generative autoregressive networks have the advantage of simplicity and have been applied for audio-driven motion generation \cite{ahn2020generative, kucherenko2020gesticulator}.
Another large branch of methods are based on variational autoencoders (VAEs), which optimize a lower bound on the data log-likelihood. 
Habibie et al. \cite{habibie2017recurrent} proposed 
a VAE with a recurrent structure for controlled motion synthesis.
In \cite{ghorbani2020probabilistic}, a hierarchical recurrent model is proposed
with each motion sub-sequence mapped to a stochastic latent code through a VAE. VAE-based methods have also found applications in cross-modal synthesis, generating motion from speech \cite{greenwood2017predicting}.

Compared to GANs and VAEs, flow-based generative models have been less explored but are gaining popularity in recent human motion generation works.
Henter et al. \cite{henter2020moglow} proposed a normalizing flow-based model called MoGlow that extend Glow \cite{kingma2018glow} to skeleton data. 
This model is adapted in \cite{alexanderson2020style} for an application of speech-driven gesture synthesis. 
Our approach is an extension to MoGlow. As a result, it features the benefits of flow-based models, which allow for tractable likelihood evaluation and efficient parameterization of encoder and decoder.
Flow-based models such as MoGlow typically process data in the Euclidean space and lack the structure for skeleton data with explicit spatial correlations. 
This may be particularly problematic when strong generalization is expected to cope with untrained uncertainties, e.g. input poses with missing markers. 
Our method remedies this by integrating graph convolutional networks.

\subsection{Pose Reconstruction from Partially Observable Data}
\label{sec:related-mm}
Reconstructing from partial observations has been a long-lasting problem and investigated in human pose estimation.
Traditional reconstruction approaches are largely based on matrix factorization. 
Peng et al. \cite{peng2015hierarchical} use adaptive non-negative matrix factorization with hierarchical blocks for motion recovery.
Wang et al. \cite{wang2016human} decompose the entire pose to partial models to exploit the abundant local body posture. Dictionary learning is designed and applied in parallel for each part. 
Gloersen and Federolf \cite{gloersen2016predicting} exploit marker inter-correlations from weighted principal components analysis (PCA) for reconstruction. 
Low-rank matrix completion can be applied to motion recovery \cite{tan2013human, hu2017motion}. 
Cui et al. \cite{cui2019nonlocal} proposed a nonlocal low-rank regularization model (NLR) utilizing kinematic information and weighted Schatten p-norm (WSN) to recover the missing markers. 
Nevertheless, these approaches make a strong assumption that perfect pose frames should be present in the sequence at least once.
In our work, this assumption is not necessary, since in each frame some markers are allowed to be missing. 

Deep neural networks, especially recurrent ones, have also been explored to encode temporal correlation for reconstructing missing markers.
Mall et al. \cite{mall2017deep} propose a deep bidirectional LSTM for denoising and synthesizing missing frames. 
Kucherenko et al. \cite{kucherenko2018neural} use similar but simpler LSTM-based and time-window-based models.
Most methods are variants of LSTM but 
suffer from the difficulty of capturing long-term dependencies. A deep bi-directional attention network (BAN) \cite{cui2019deep} is proposed and embedded in the bidirectional LSTM to alleviate this issue. 
Lohit and Anirudh \cite{lohit2021recovering} consider the problem of reconstructing completely unobserved markers of motion sequences. The reconstruction 
is solved by projecting the observed action onto the action manifold via latent space optimization. 
Similar to previous approaches, we also use an LSTM to account for temporal causalities.
Furthermore, our method encodes spatial and temporal relations between markers and input frames with graph convolutional networks.

\subsection{Graph-based Human Body Representation}
\label{sec:related-gb}
Graph neural networks (GNNs) have received increasing attention and have been successfully applied to represent human skeleton data.
Most works focus on discriminative tasks such as skeleton-based action recognition
\cite{yan2018spatial, li2019actional, cheng2020skeleton} 
and group behavior recognition \cite{yang2020group}. 
Si et al. \cite{si2019attention} employ a graph convolutional LSTM network with an attention mechanism to enhance information extraction. 
Inspired by deformable part-based models (DPMs), \cite{thakkar2018part} divides the skeleton graph into subgraphs and proposes a part-based graph convolutional network (PB-GCN) for action recognition. 
In \cite{wen2019graph}, motif-based graph convolution is proposed to learn hierarchical spatial structures for action recognition. 

For human motion prediction and generation, 
Li et al. \cite{li2020dynamic} propose a multi-scale graph computational unit (MGCU) to extract deep features. 
Jain et al. \cite{jain2016structural} construct a structural graph in which the nodes and edges consist of LSTMs to model the body dynamics. 
Yan et al. \cite{yan2019convolutional} design a graph-based framework called convolutional sequence generation network (CSGN) to directly generate the entire sequence instead of sequentially. 
In addition, graph convolution networks have been applied to pose regression \cite{zhao2019semantic}, trajectory generation \cite{yang2020impact}, and human video prediction \cite{zhao2020pose}. 
In \cite{liu2019graph}, the framework of normalizing flows is extended with a graph auto-encoder to generate non-human graph structure samples.  
Inspired by the ST-GCN \cite{yan2018spatial}, graph normalizing flows (GNFs) \cite{liu2019graph} and MoGlow \cite{henter2020moglow}, we use spatial graph convolutional network (S-GCN) to encode the spatial relationship of the human skeleton in normalizing flows and leverage ST-GCN to extract features from past sequences.

\section{METHOD} \label{sec:method}
This section formulates our target problem and establishes notations used throughout the paper. Preliminaries about normalizing flow, MoGlow and spatial-temporal graph convolutional networks are also given. On the basis of these, we present the contributed framework.

\subsection{Problem Formulation}
Suppose that a 3D skeleton-based pose at time $t$ is denoted as
$X^{(t)}\in \mathbb{R}^{M \times C}$, with $M$ joint markers and $C=3$ feature dimensions. 
The past human poses till time step $t_0$ are  $\mathbb{X}_{(t_0-T_h):(t_0-1)}=[X^{(t_0-T_h)},...,X^{(t_0-1)}] \in \mathbb{R}^{M \times C \times T_h}$.
The goal is to estimate a parameterized probabilistic model $p_{\theta}$ from a set of human pose trajectory data, where the optimal parameters are given by:
\begin{equation}
    \theta^* = \operatornamewithlimits{argmax}_{\theta} p_{\theta}( \mathbb{X}_{(t_0):(T)} | \mathbb{X}_{(t_0-T_h):(t_0-1)}, C_{(t_0-T_h):(T)} )
\end{equation}
from which one can sample to predict $T+1$ future poses $\mathbb{X}_{(t_0):(T)}$, given the past poses $\mathbb{X}_{(t_0-T_h):(t_0-1)}$ and a control input $C_{(t_0-T_h):(T)}$ for the full sequence.
%
Sometimes, the past poses are only partially observed, e.g. with frames missing some marker positions. 
In such cases, the task becomes that of reconstructing a full trajectory $\mathbb{X}_{(t_0):(T)}$ from an imperfect input, an input that we denote $\hat{\mathbb{X}}_{(t_0-T_h):(t_0-1)}$.
%

\subsection{Normalizing Flows and MoGlow}
\label{sec:normflow}
Normalizing flows \cite{dinh2014nice, dinh2016density, kingma2018glow} are a class of generative models that allow efficient sampling and inference. 
The idea is to find an invertible transformation $z=f(X)$ with inverse $X=f^{-1}(f(X))$ 
to map the data $X$ into a latent space where the distribution is tractable, such as a multivariate Gaussian $p_\theta(z)$. 
For a given distribution of $z$, the change-of-variable rule gives
%
\begin{equation}
    p(X) = p(z)\lvert \textrm{det}\frac{\partial f(X)}{\partial X} \rvert ,
\end{equation}
where $\frac{\partial f(X)}{\partial X}$ is the Jacobian matrix of $f^{-1}$ at $X$. 
To obtain a complex distribution with expressive mapping, a series of invertible transformations are chained together. The relationship between input $X$ and latent representation $z$ becomes: 
\begin{equation}
    \label{eq:f}
    X \xleftrightarrow{f_1} h_1 \xleftrightarrow{f_2} h_2 \cdots \xleftrightarrow{f_K} h_K, ~ z \triangleq h_K
\end{equation}
The sequence of invertible transformations $f^{-1}$ in Equation \ref{eq:f} is known as a normalizing flow. 
We can generate samples $X$ by first sampling $z\sim p_\theta(z)$ and then computing $X=f^{-1}(z)$. 
Then, we can write the log-likelihood of $X$ as
\begin{align}
    \log p_{\theta}(X)  &= \log p_{\theta}(z) +  \log \lvert \textrm{det}\frac{\partial z}{\partial X} \rvert \\
    &=  \log p_{\theta}(z) + \sum_{i=1}^{K} \log \lvert \textrm{det} \frac{\partial h_i}{\partial h_{i-1}} \rvert
\end{align}
To get a tractable determinant of the Jacobian matrix, the idea is to select the transformation whose Jacobian is a triangular matrix. The log-determinant is then simplified as 
\begin{equation}
     \log \lvert \textrm{det} \frac{\partial h_i}{\partial h_{i-1}} \rvert = \textrm{sum}( \log \lvert \textrm{diag} (\frac{\partial h_i}{\partial h_{i-1}})\rvert ).
\end{equation}
%

Glow \cite{kingma2018glow} is a normalizing flow-based generative model, which has achieved impressive performance for facial image synthesis.
In Glow, each step of flow consists of three sub-steps: \textit{actnorm}, \textit{invertible 1$\times$1 convolution}, and an \textit{affine coupling layer}. 
Actnorm is an activation normalization layer that applies a scale and bias with data-dependent initialization.
1$\times$1 convolution is a linear transformation layer for soft permutation. 
The affine coupling layer splits the input into two parts. One half of the input can be affinely transformed based the other half. The other half is kept unchanged, which leads to an easy reverse transformation.

MoGlow \cite{henter2020moglow} extends Glow to skeleton sequences by describing the distribution of future poses in terms of an autoregressive model.
It adds control signals to achieve control over the output and uses a recurrent neural network, an LSTM, 
to integrate information over time.
The past and current control signal $C_{(t-T_h):(t)}$, and past human poses $\mathbb{X}_{(t-T_h):(t-1)}$ are conditioning information. 
MoGlow simply concatenates the current pose with additional conditioning information and feeds it into the LSTM
in the affine coupling layers. 
The autoregressive model can be written as 
\begin{align}
    p_{\theta}(\mathbb{X}|C) = & p(\mathbb{X}_{(t_0-T_h):(t_0-1)}|C_{(t_0-T_h):(t_0-1)})\\
      &\cdot \prod_{t=t_0}^{T}p_\theta(X_t|\mathbb{X}_{(t-T_h):(t-1)}, C_{(t-T_h):(t)}, H_{t-1}),
\end{align}
where $H_t$ is the latent state of the LSTM. 
In our proposed model, indicative features of the past poses are first extracted using spatial-temporal graph convolutional networks before being fed into the recurrent neural network, which will be explained in the following. 

\subsection{Graph Motion Glow}

\begin{figure*}[htp]
  \centering
  \includegraphics[width=1\textwidth]{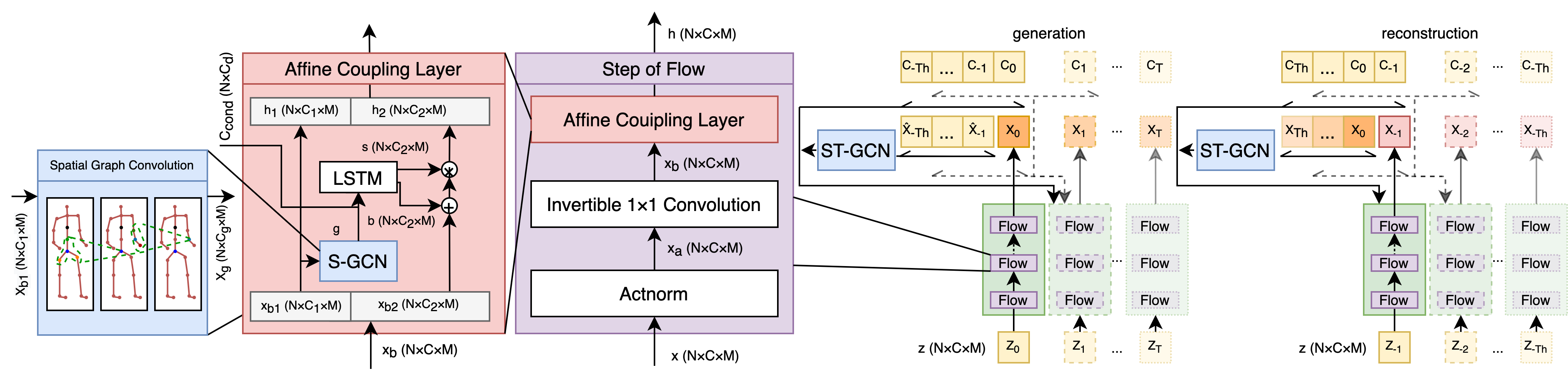}
  \caption{
  Overview of the Graph Motion Glow for skeleton-based motion reconstruction and generation.
  The body markers are connected as a skeleton graph and fed into the step of flows the with control input and history input. 
The S-GCN in the affine transformation encodes the spatial relationships among markers,
and the LSTM preserves the temporal information. 
The ST-GCN extracts features from past sequences. 
}
  \label{fig:overview-framework}
\end{figure*}

\begin{figure}[ht]
  \centering
  \includegraphics[width=0.4\textwidth]{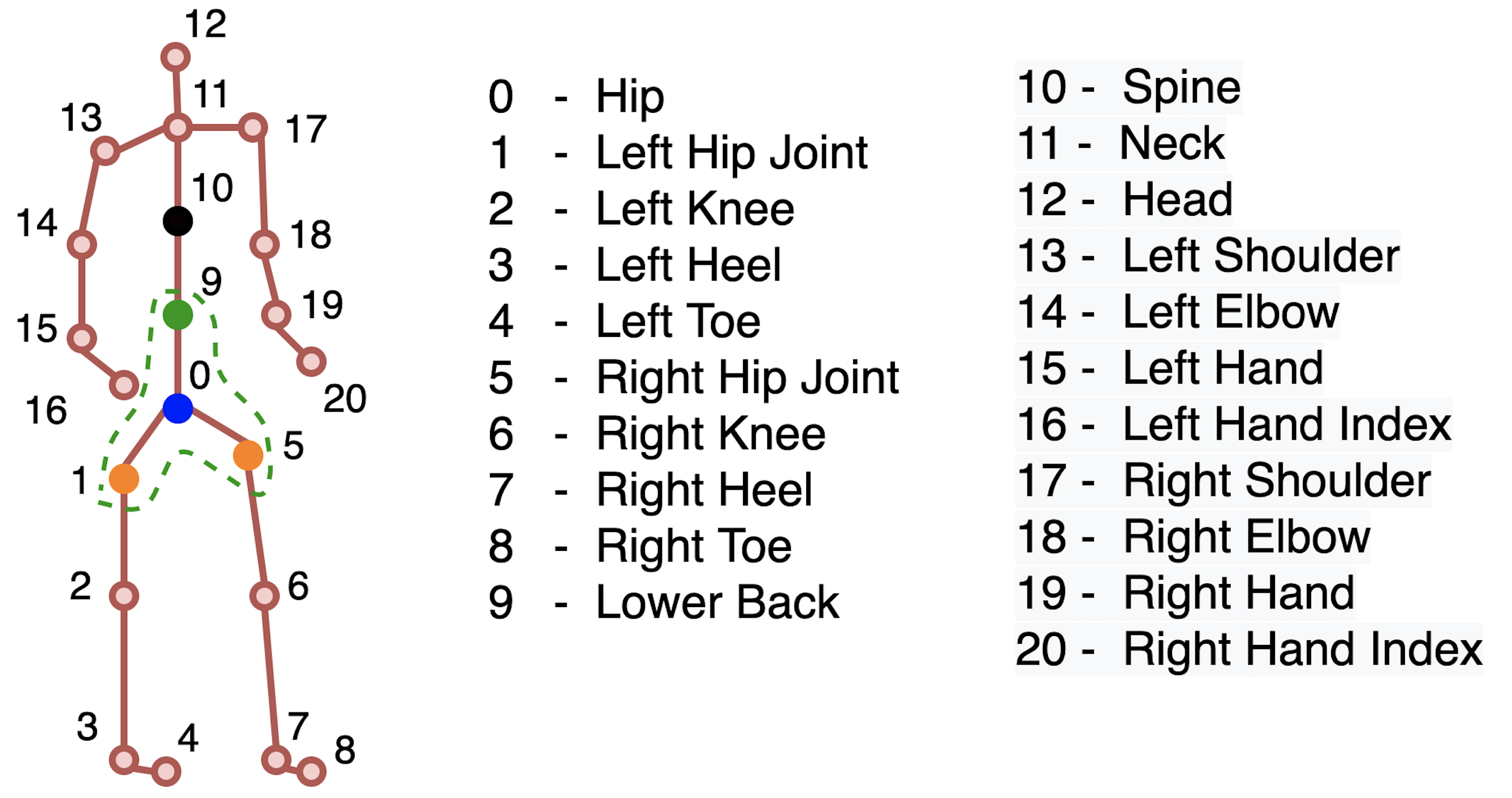}
  \caption{The spatial graph of a skeleton used in this work. 
    Each graph node corresponds to the marker on the right~\cite{ikhansul17-vaelstm}. 
    The edges between nodes are defined based on natural joints relationships. 
  }
  \label{fig:skeleton-graph}
\end{figure}

In MoGlow, the coordinates of the skeleton data are concatenated to one single feature vector per frame. 
With a spatial graph neural network (S-GCN), 
we instead convert the skeleton data into an undirected graph $G=(V, E)$, where $V$ is the set of nodes and $E$ is the set of edges, based on the natural spatial relationships between parts of the skeleton. 
Each joint marker is represented by a node and the bone between two connected markers by an edge. 
The skeleton graph used in this work is illustrated in Fig. \ref{fig:skeleton-graph}. 
%

Spatial graph convolutional networks extend convolutions, typically applied to images, to graph structures \cite{dai2017deformable}. 
In the case of images, pixels within a rectangular neighborhood have a fixed spatial order, but no such order exists in a skeleton graph. 
To address this problem, 
Yan et al. proposed a spatial configuration partition strategy \cite{yan2018spatial}.
We follow the same idea to define a sampling function for convolution. 
The spatial configuration partition strategy divides each neighborhood of a node into three subsets, i.e., the neighbors located closer to the graph center, the neighbors that are farther away, and the node itself. 

As shown in  Fig. \ref{fig:skeleton-graph}, we assume the chest node, marker-10, to be the center of the skeleton. 
For the neighbor set of marker-0 with kernel scale $D=1$ (nodes within the green dotted line), there are three subsets: marker-9 (node closer to the center, green), marker-1 and marker-5 (nodes farther away, orange), and marker-0 (the node itself, blue). 
Similar neighborhood subsets are determined for each node in the skeleton graph for different kernel scales.

A spatial graph convolution can then be defined as:
\begin{equation}
    y_i=\sum_{v_j \in S_i} \frac{x_j}{D_{v_i}(v_j)} w(l_{v_i}(v_j)),
\end{equation}
where $x_i$ is the feature vector of node $v_i$ before the convolution, $y_i$ is the corresponding vector after the convolution, $S_i$ is the neighbor set of $v_i$, and $w$ is a weight function that depends on
 the spatial partition strategy $l_{v_i}(v_j)$ that maps each node $v_j$ to its corresponding subset. 
$D_{v_i}(v_j)$ is the number of nodes in the subset, which is used as a normalizing term to compute an average per subset. 

We also employ temporal connections that connect the same node in consecutive frames with temporal edges. 
For a past sequence,
$v_{t,i}$, at time $t$, connects to $v_{t-1,i}$ and $v_{t+1,i}$ along the temporal dimension. 
The whole graph sequence is thus composed of a spatial graph and a temporal graph. 
As in ST-GCN \cite{yan2018spatial}. each layer of the network includes one S-GCN followed by a temporal convolution network (TCN) over the temporal domain, with residual connections. 

\begin{figure*}[ht]
\begin{minipage}[t]{1\textwidth}
\centering
\includegraphics[width=14.3cm]{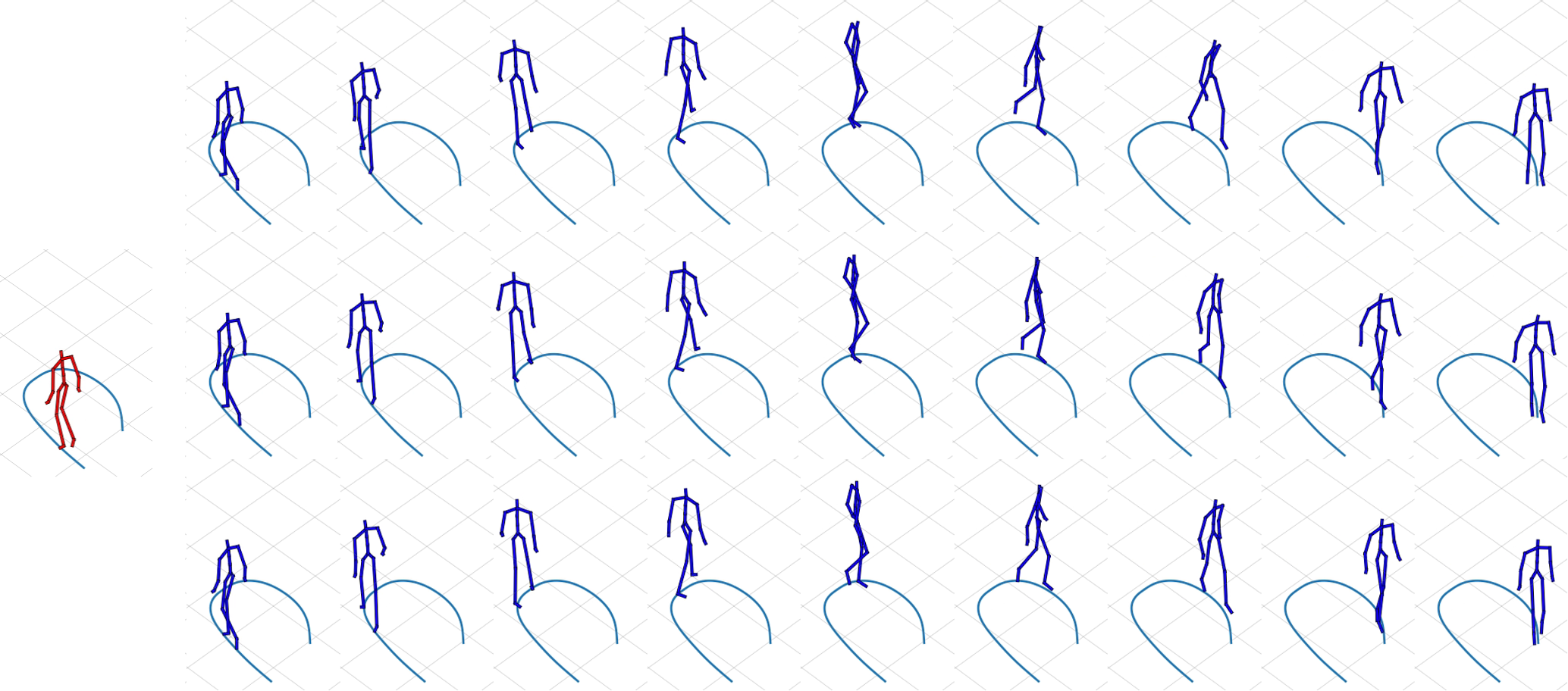}
\end{minipage}
\begin{minipage}[t]{1\textwidth}
\centering
\includegraphics[width=14.3cm]{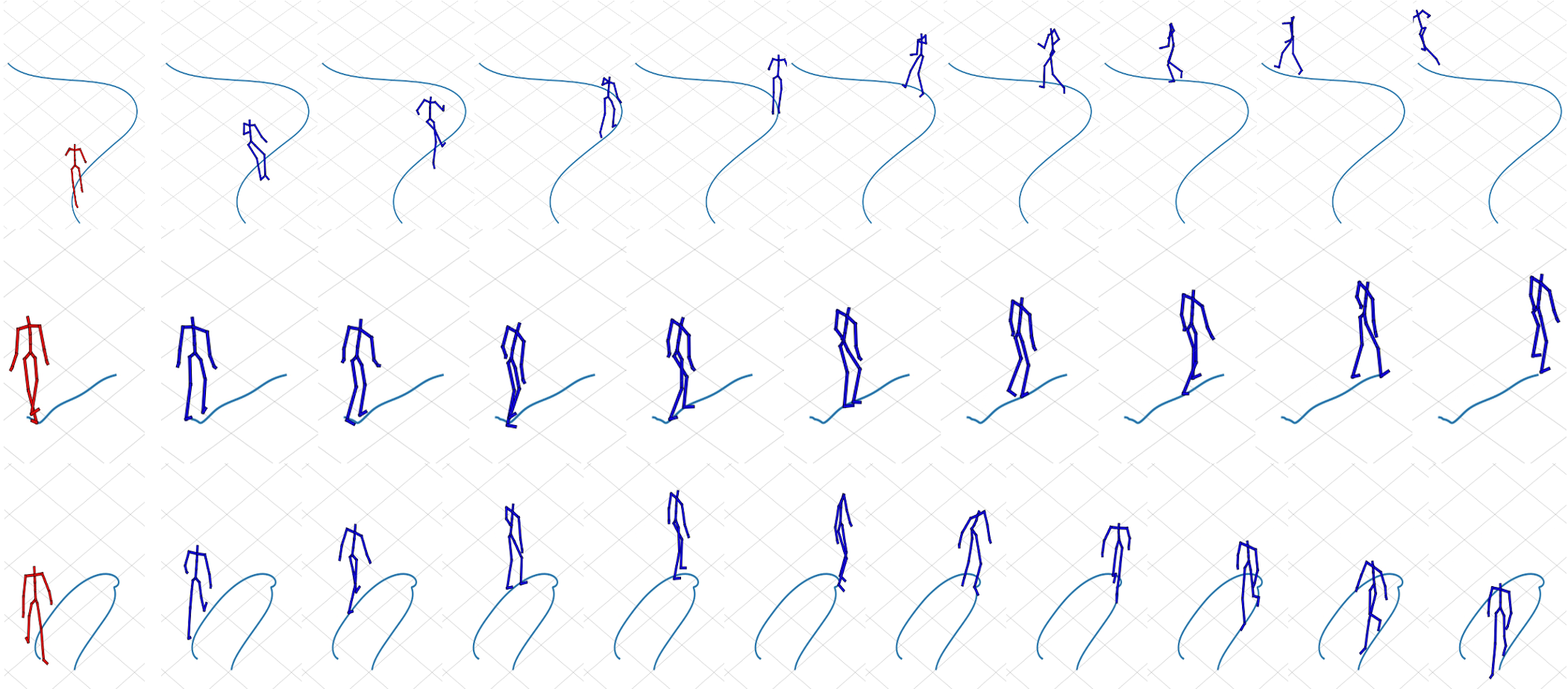}
\end{minipage}
\caption{
Example sequences generated by STMG. 
The top three sequences are generated from the same past information. 
The bottom three sequences are generated given different past information. 
}
\label{fig:exp-diversity}
\end{figure*}

\begin{figure*}[ht]
  \centering
  \includegraphics[width=14.3cm]{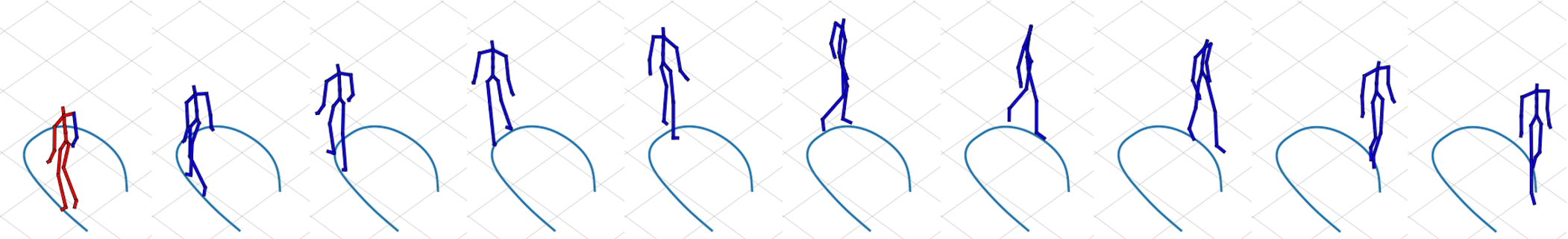}
  \caption{ 
    Example sequence generated and reconstructed by STMG.
    In the past poses, the same as the first three rows of Fig. \ref{fig:exp-diversity}, the markers of the right arm are set to zero. 
  }
  \label{fig:37}
\end{figure*}

An overview of the proposed graph motion glow framework is presented in Fig. \ref{fig:overview-framework}, which includes each component transformation. 
Our flow model includes the three main reversible transformation layers in Glow, but  
extended to graph structures. 
We further use ST-GCN to extract features from the autoregressive history input. 
The input $X$ and output $z$ are represented as tensors of shape $[M\times C\times T_h]$ with spatial dimension $M$, channel dimension $C$ and temporal dimension $T_h$. 
$X_a$ and $X_b$ denote intermediate results of the actnorm layer and invertible 1$\times$1 convolution layer, which performs a soft permutation of channels.
The affine coupling layer is more complex. 
The idea is to split the input channel-wise into two parts and transform one part based on the other part and conditioning information. 
Mathematically, we define the input $X_b$ and output  $h$ of the affine coupling layer as concatenations $X_b=[X_{b1}, X_{b2}]$ and $h=[h_{1}, h_{2}]$.
The coupling can then be written 
\begin{equation}
    [h_{1}, h_{2}]=[X_{b1}, (X_{b2}+\textbf{b})\odot \textbf{s}],
\end{equation}
where $\odot$ is a Hadamard product and the scaling $\textbf{s}$ and bias $\textbf{b}$ terms are computed with S-GCN, ST-GCN and LSTM:
%
\begin{align}
    g_t &= SGCN(X_{b1, t}),\\
    p_t &= STGCN(\hat{X}_{(t-T_h):(t-1)}),\\
    [\textbf{s}_t, \textbf{b}_t]&=LSTM(g_t, p_t,C_{(t-T_h):(t)}).
\end{align}

Here, S-GCN captures the spatial graph information $g_t$ from markers in the current time step $t$, 
ST-GCN extracts spatial-temporal features $p_t$ from the past frames of the sequence, 
and LSTM produces the scaling and bias with dependencies over time.  
At different steps of the flow, we use spatial graph convolutions with different graph kernel scales $D$ to capture the hierarchical structure of the body.

For generation, we exploit the fact that the flow model is reversible.
Thus we can generate a new pose $X_t$ from the trained graph motion glow model using a latent vector $z_t$ (see Section \ref{sec:normflow}), the recent history of poses $\hat{\mathbb{X}}_{(t-T_h):(t-1)}$ and control input $C_{(t-T_h):(t)}$ for the full sequence. 
The latent vector $z_t$ is sampled independently from a standard Gaussian distribution.
The generated $X_t$ then becomes a part of the conditioning information for generating the next following pose $X_{t+1}$. 
%
During training, data was augmented by lateral mirroring and time-reversion. This allows an imperfect input with missing markers to be reconstructed with the same framework by reversing the generated sequences $\mathbb{X}_{(t_0):(T_h)}$ and control signal $C_{(t_0-T_h):(T_h)}$ to $\mathbb{X}_{(t_h):(T_0)}$ and $C_{(T_h):(t_0-T_h)}$. 
With these reversed sequences regarded as control information, an imperfect input can be reconstructed with generated markers now used to fill in the holes of the missing data.

\section{EXPERIMENTAL RESULTS}
\label{sec:experiment}

In this section, we evaluate the performance of our model. 
We first describe the dataset and experiment setup, and then continue with the qualitative and quantitative results on generating and reconstructing human motion samples of high fidelity. On basis of these, we discuss strengths and limitations of the baseline and proposed frameworks.

\subsection{Dataset}

We consider a human locomotion dataset preprocessed by \cite{henter2020moglow}, which pools samples from the Edinburgh Locomotion MOCAP\cite{ikhansul17-vaelstm}, CMU Motion Capture \cite{cmu}, and HDM05 \cite{MuellerBS09_MocapAnnotation_SCA} datasets. 
The dataset is downsampled to 20 fps and sliced into fixed-length sequences of 80 frames with 50\% overlap for training and augmented by lateral mirroring and time-reversal. 
Each data instance resembles a clip of a human locomotion animation, i.e., movement including various gaits along different curved paths. 

The motion is represented by the 3D coordinates of 21 joints at each frame, as displayed in Figure \ref{fig:skeleton-graph}. 
In addition, there are 3 scalar control signals indicating forward, sideways and rotational velocities for each frame.
To generate incomplete MoCap frames, we set some markers to zero with a binary matrix $M_b$ in the past poses used for generation: $\hat{\mathbb{X}}_{(t_0-T_h):(t_0-1)}=M_b \odot \mathbb{X}_{(t_0-T_h):(t_0-1)}$.
We preprocess these sequences with the following five settings: 
a)~remove the markers on the right arm (markers 18, 19, 20), 
b)~remove the markers on the left leg (markers 2, 3, 4), 
c)~remove the markers on the right arm and left leg, 
d)~randomly remove 4 markers,
e)~keep all markers intact.

\subsection{Proposed Model and Ablations}
\label{sec:expmodel}
The proposed graph motion glow model is trained with a 10-frame time window, similarly to MoGlow. 
In our experiments, all glow models were structured with 16 steps of flows. The structure of one step of flow is illustrated as the purple block in Fig. \ref{fig:overview-framework}. 
Each affine coupling step consisted of a S-GCN layer and an LSTM with two layers, shown as the red block in Fig. \ref{fig:overview-framework}. Note that ST-GCN is applied to the past history of frames with the same features used in each such step.
The sizes of graph kernels for the S-GCN are 3, 5, and 7 for the first 10 flow steps, the following 4 steps, and the last 2 steps.
We use a temporal kernel with a size of 9 and 512 hidden cells for each LSTM layer. 

To assess the impact of design decisions, we trained three versions of the architecture on the human data. 
Each version has specific components disabled to examine their effects comparing with our full framework. 
We denote our proposed graph-based model, the spatial-temporal graph motion glow as "STMG". 
The first ablated configuration "SMG" only uses the spatial graph convolution networks without temporal convolution, i.e., the temporal convolution in the "ST-GCN" block in Fig. \ref{fig:overview-framework} is turned off. 
The second configuration uses no graph structure and is equivalent to the MoGlow baseline MoGlow, denoted as "MG". 

\subsection{Results and Discussions}

We present examples of generated sequences in Fig. \ref{fig:exp-diversity}, which shows snapshots of every 10 frame from sequences of length 100. A video with generated examples can be found at \href{https://kth.box.com/s/2vngw2tu1pg217cf9bo4s98fzobnccxb}{this link}\footnotemark[4].
The results demonstrate the diversity and quality of samples. 
The top three sequences are sampled from the same 10-frame seeding history but show different poses, which means the probabilistic model does not collapse into a stereotypical mode. 
The bottom three sequences are generated given different past poses and control inputs, which show our model incorporating spatial and temporal graph structure can generate long-term locomotion behavior of high fidelity and diversity. 
When the past information is incomplete, we further reconstructed the missing markers. 
In Fig. \ref{fig:37}, we observe that the reconstructed markers (the right arm in the first frame of Fig. \ref{fig:37}) fit the original skeleton well.
We further evaluate the quality of generated and reconstructed motions by performing both footsteps analysis (see \cite{henter2020moglow} for detailed definitions)
and bone-length analysis. 

\footnotetext[4]{\url{https://kth.box.com/s/2vngw2tu1pg217cf9bo4s98fzobnccxb}}

\subsubsection{Footstep Analysis}
Footsteps analysis is used to evaluate foot-sliding artifacts in locomotion synthesis. 
In footstep analysis, footsteps are detected as time intervals when the horizontal speed of the heel joints is below a tolerance value $v_{tol}$.
Because of foot-sliding artifacts, the heels sometimes exhibit a spurious behavior with the horizonal speed exceeding a specified tolerance value. 
We incremented the tolerance $v_{tol}$ in small steps. The total number of detected footsteps first rises, then reaches the maximum value. 
The number of identified footsteps decreases if the tolerance increasing further. 
Motion sequences with larger foot-sliding artifacts need a higher tolerance value to reach the maximum value of estimated footsteps. 

\begin{figure}[htp]
  \centering
  \vspace{-12pt}
  \includegraphics[width=0.38\textwidth]{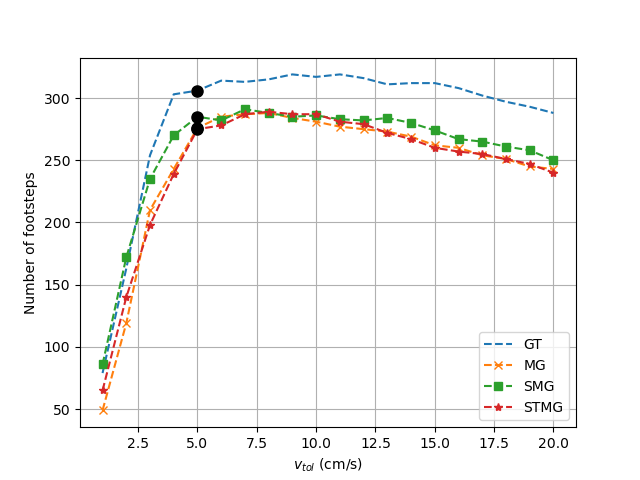}
  \caption{ 
    Footstep analysis for samples generated with complete past input: footstep count $f_{est}$ on tolerance value $v_{tol}$.
    Black dots incidate the location of $v_{tol}^{95}$.
  }
  \label{fig:full}
\end{figure}

\begin{figure}[ht]
\vspace{-10pt}
\subfloat[]{
\begin{minipage}[t]{0.24\textwidth}
\centering
\vspace{-10pt}
\includegraphics[width=4.3cm]{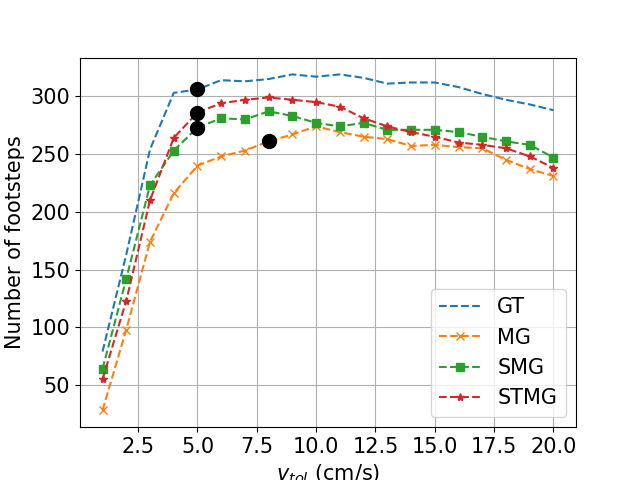}
\end{minipage}
}
\subfloat[]{
\begin{minipage}[ht]{0.24\textwidth}
\centering
\includegraphics[width=4.3cm]{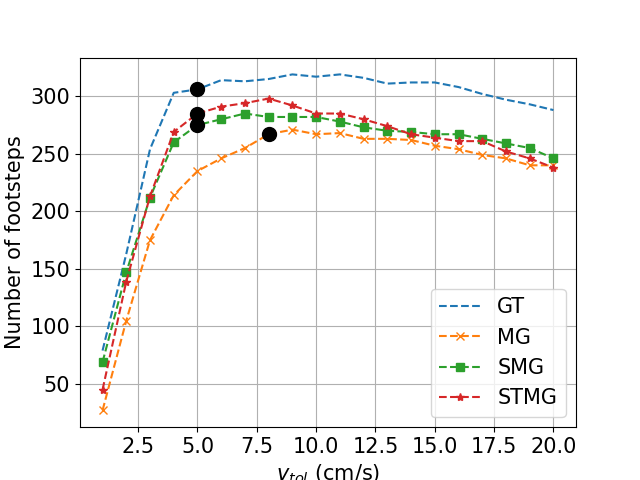}
\end{minipage}

}
\vspace{10pt}
\\
\subfloat[]{
\begin{minipage}[ht]{0.24\textwidth}
\centering
\includegraphics[width=4.3cm]{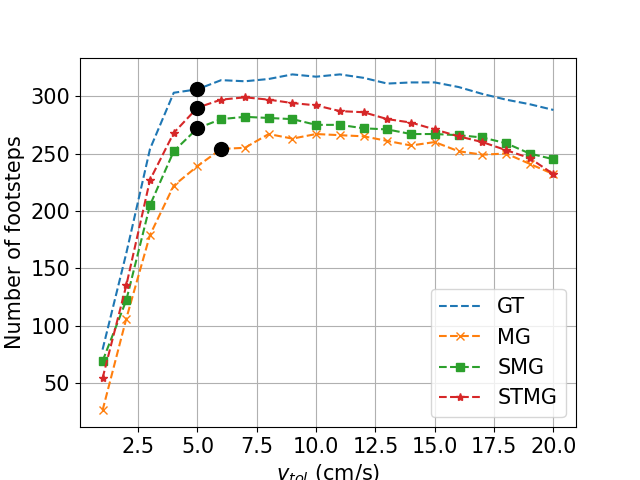}
\end{minipage}
}
\subfloat[]{
\begin{minipage}[ht]{0.24\textwidth}
\centering
\includegraphics[width=4.3cm]{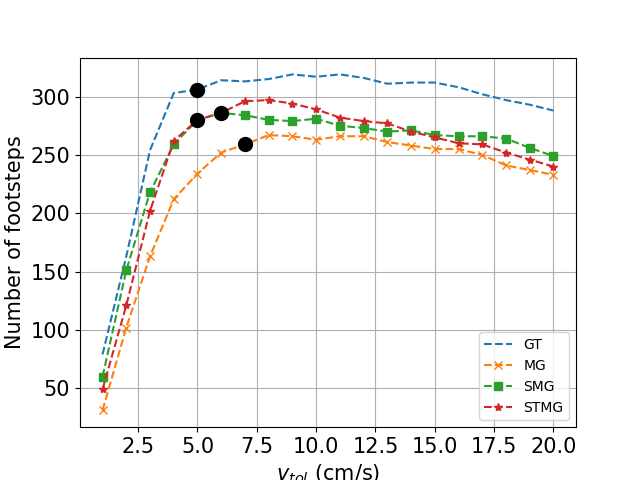}
\end{minipage}
}\\
\caption{Footstep analysis for incomplete past input: footstep count $f_{est}$ for each tolerance $v_{tol}$. Black dots indicate $v_{tol}^{95}$. Graphs represents cases with missing markers (a)~on the right arm (marker-18, 19, 20),  
    (b)~on the left leg (marker 2, 3, 4), 
    (c)~on the right arm and left leg, and
    (d)~4 randomly missing.}
\label{fig:miss}
\end{figure}

The estimated number of footsteps for different tolerance values in generated motions are shown in Fig. \ref{fig:full} and Fig. \ref{fig:miss}. Each results is based on an average over 150 generated sequences.
For evaluation, we detect the first tolerance value $v_{tol}^{95}$, for which at least 95\% of the maximum number of footsteps are estimated. These values are shown as black dots in the figures. 
The total estimated number of footsteps, speed tolerance for capturing 95\% steps, and the mean and standard deviation of the step duration are shown in Table \ref{tab:ft}. 
We note that in Fig. \ref{fig:full}, 
which assumes sequences without missing markers,
the curves are close.
The performance of the proposed method is comparable to the state-of-the-art baseline.
However, when the given past data is incomplete, i.e., some markers are missing, we note in Fig. \ref{fig:miss} that the curves of our proposed spatial temporal graph motion glow model are consistently closer to the curve of ground truth, 
illustrating the improvement in robustness of the proposed graph model.
%
We also observe from the ablation study that STMG outperforms SMG, showing an extra performance gain from the temporal convolutions of the past frames.

\begin{table}[ht]
\centering
\begin{tabular}{c|c|cccc}
\hline
Miss                                                                  & Model & $f_{est}$ & $v_{tol}^{95}$ & $\mu$          & $\sigma$       \\ \hline
-                                                                     & GT    & 5         & 306            & 0.315          & 0.273          \\ \hline
\multirow{3}{*}{-}                                                    & MG    & 5         & 276            & 0.298          & 0.318          \\
                                                                      & SMG   & 5         & \textbf{285}   & 0.294          & 0.242          \\
                                                                      & STMG  & 5         & 275            & \textbf{0.316} & \textbf{0.267} \\ \hline
\multirow{3}{*}{RA}                                                   & MG    & 8         & 261            & 0.380           & \textbf{0.283} \\
                                                                      & SMG   & 5         & 273            & 0.281          & 0.231          \\
                                                                      & STMG  & 5         & \textbf{286}   & \textbf{0.302} & 0.249          \\ \hline
\multirow{3}{*}{LL}                                                   & MG    & 8         & 267            & 0.382          & 0.362          \\
                                                                      & SMG   & 5         & 275            & 0.287          & 0.248          \\
                                                                      & STMG  & 5         & \textbf{285}   & \textbf{0.314} & \textbf{0.275} \\ \hline
\multirow{3}{*}{\begin{tabular}[c]{@{}c@{}}RA\\ \&\\ LL\end{tabular}} & MG    & 6         & 254            & 0.327          & 0.318          \\
                                                                      & SMG   & 5         & 272            & 0.275          & 0.217          \\
                                                                      & STMG  & 5         & \textbf{290}   & \textbf{0.306} & \textbf{0.267} \\ \hline
\multirow{3}{*}{R4M}                                                  & MG    & 7         & 256            & 0.357          & 0.329          \\
                                                                      & SMG   & 5         & 280            & 0.307          & 0.252          \\
                                                                      & STMG  & 6         & \textbf{286}   & \textbf{0.315} & \textbf{0.253} \\ \hline
\end{tabular}
\caption{Results of foot-step analysis for motion generation: total number of footsteps $f_{est}$, speed tolerance for capturing 95\% steps $v_{tol}^{95}$, mean $\mu$ and standard deviation $\sigma$ of step-duration. We remove a few markers in the past poses. RA: right arm; LL: left leg; R4M: random 4 markers. The numbers closest to the ground truth are shown in bold.}
\label{tab:ft}
\end{table}

\begin{table}[ht]
\centering
\begin{tabular}{c|c|cccc}
\hline
\multirow{2}{*}{Miss}                                                 & \multirow{2}{*}{Model} & \multicolumn{2}{c}{Generation}  & \multicolumn{2}{c}{Reconstruction} \\ \cline{3-6} 
                                                                      &                        & RMSE           & $\sigma$       & RMSE             & $\sigma$        \\ \hline
\multirow{3}{*}{-}                                                    & MG                     & 0.597          & 0.067          & -                & -               \\
                                                                      & SMG                    & \textbf{0.191} & \textbf{0.039} & -                & -               \\
                                                                      & STMG                   & 0.779          & 0.073          & -                & -               \\ \hline
\multirow{3}{*}{RA}                                                   & MG                     & 279301         & 7.532          & 137325           & 4.471           \\
                                                                      & SMG                    & \textbf{0.842} & \textbf{0.051} & 2.589            & 0.075           \\
                                                                      & STMG                   & 0.881          & 0.079          & \textbf{0.890}   & \textbf{0.059}  \\ \hline
\multirow{3}{*}{LL}                                                   & MG                     & 834081         & 12.558         & 690231           & 7.663           \\
                                                                      & SMG                    & 1.935          & \textbf{0.050} & 3.701            & 0.130           \\
                                                                      & STMG                   & \textbf{0.909} & 0.080          & \textbf{2.336}   & \textbf{0.122}  \\ \hline
\multirow{3}{*}{\begin{tabular}[c]{@{}c@{}}RA\\ \&\\ LL\end{tabular}} & MG                     & 537036         & 8.788          & 2404434          & 7.533           \\
                                                                      & SMG                    & 2.134          & \textbf{0.052} & 3.687            & 0.103           \\
                                                                      & STMG                   & \textbf{0.917} & 0.080          & \textbf{1.864}   & \textbf{0.093}  \\ \hline
\multirow{3}{*}{R4M}                                                  & MG                     & 787638         & 12.138         & 80931            & 2.897           \\
                                                                      & SMG                    & \textbf{0.542} & \textbf{0.044} & 6.589            & 0.092           \\
                                                                      & STMG                   & 0.938          & 0.080          & \textbf{1.072}   & \textbf{0.065}  \\ \hline
\end{tabular}
\caption{Results of bone-length analysis for human motion generation and reconstruction. The best values are in bold.}
\label{tab:bl}
\end{table}

\subsubsection{Bone-length Analysis}
The human skeleton is represented by joint coordinates, without explicit consideration of spatial constraints such as the length of bones. To evaluate the data quality on this aspect, \cite{henter2020moglow}  performs bone-length analysis to detect artifacts such as flying-apart joints.
The analysis looks at the bone-length Root Mean Squared Error (RMSE) $bl_{rmse}$ ($cm$) and standard deviation $bl_{\sigma}$ ($cm^2$), which represent the consistency, stability and quality of generation and reconstruction. 
%
Table \ref{tab:bl} reports the results on the bone-length analysis. 
Given complete past frames, all competing models achieve relative small RMSE and $\sigma$, with SMG outperforming the other two.

Again, the performance of baseline and our approaches differs significantly when the models are tasked to generalize from untrained imperfect data. MG performs poorly on this indicator, exhibiting huge bone-length artifacts.
In contrast, for both SMG and STMG, the RMSE and $\sigma$ of bone-lengths are still small. 

\begin{figure}[htp]
\vspace{-5pt}
\subfloat[]{
\begin{minipage}[t]{0.22\textwidth}
\centering
\includegraphics[width=2.5cm]{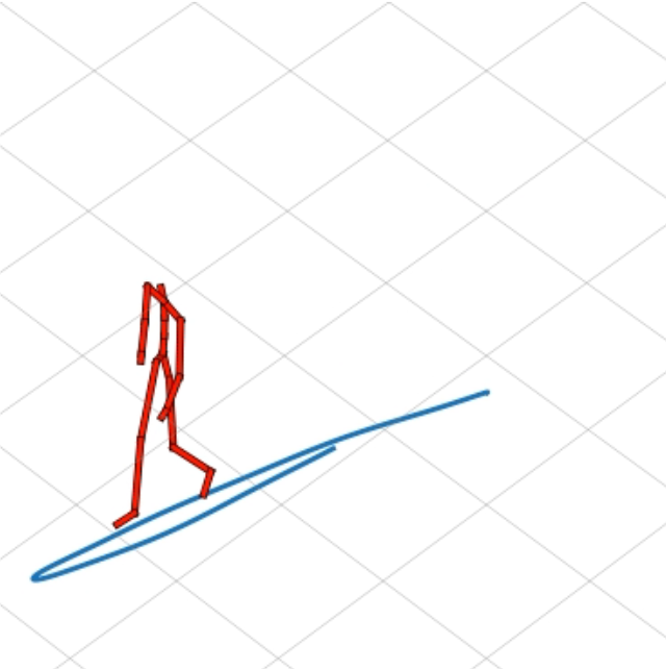}
\end{minipage}
}
\subfloat[]{
\begin{minipage}[t]{0.22\textwidth}
\centering
\includegraphics[width=2.5cm]{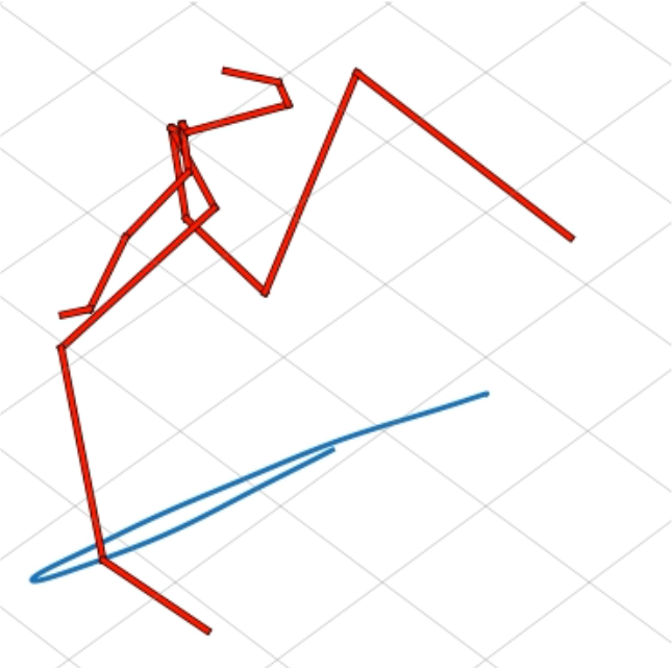}
\end{minipage}
}
\\
\caption{Examples of pose generated by MG. (a)~with complete past poses. (b)~with incomplete past poses. }
\label{fig:m27-147}
\end{figure}

A closer look at the samples reveals unstable generation and reconstruction with MG. A typical example is shown in Fig. \ref{fig:m27-147}. When the conditioning information includes complete past poses, the generated pose is stable. 
However, when the conditioning information is incomplete, with four random markers missing, we observe that some joints fly apart. This situation occurs occasionally. With incomplete conditioning information, there exists a distribution shift that may result in bone-stretching or joints flying apart, since the feed-forward neural networks are sensitive to the shifts of the data domain. 
For SMG and STMG, the generated poses are always stable and the RMSE and $\sigma$ of bone lengths are small, since we exploit the correlation of human skeletons to enforce the bias with graph structures. 

\section{Conclusion}
\label{sec:result}
In this paper, we propose a graph-based normalizing~flow model to tackle the problem of human motion generation and reconstruction. 
This new modelling framework has the following main advantages:
(1)~It is an extension of MoGlow that is probabilistic and allows inference of the exact likelihood. 
(2)~It utilizes spatial-temporal graph convolutional networks to improve the robustness of generation. 
To the authors' knowledge, this is the first work where the graph-based normalizing flow is used to generate and reconstruct human motion. From the results is can be concluded that it overcomes some of the limitations in earlier models. 
In the future, we plan to extend the graph-based motion glow model to multiple scales to tackle more complex motions. 
\section*{Acknowledgements}
This research has received funding from the EC Horizon 2020 research and innovation program under grant agreement n. 824160 (EnTimeMent). 
\bibliographystyle{IEEEtran}
\balance
\bibliography{main}
\end{document}